\documentclass[letterpaper,10pt,conference]{ieeeconf}

\IEEEoverridecommandlockouts
\usepackage{graphicx}
\usepackage{amssymb}
\usepackage{amsmath}
\usepackage{booktabs}
\usepackage[table]{xcolor}
\definecolor{groupgray}{RGB}{238,238,238}
\definecolor{bestyellow}{RGB}{255,249,210}
\definecolor{oursgreen}{RGB}{222,240,216}
\definecolor{suturebotheader}{RGB}{238,250,238}
\definecolor{srthheader}{RGB}{241,241,253}

\title{\LARGE \bf
SurgCast: Action-Conditioned Future Skeletons for Controllable Surgical World Models
}

\author{%
\authorblockN{\parbox{0.98\textwidth}{\centering
Wanhao~Liu\textsuperscript{2,3,*},
Rulin~Zhou\textsuperscript{1,*},
Liangjing~Shao\textsuperscript{1,*},
Zhaocheng~Lin\textsuperscript{3},
Dongyue~Li\textsuperscript{3},
Jinsong~Lin\textsuperscript{1},
Guoheng~Ma\textsuperscript{2},
Hangheng~Yang\textsuperscript{3},
Zhiqing~Tang\textsuperscript{1},
Jing~Chen\textsuperscript{3},
Panshuo~Li\textsuperscript{3},
Hongliang~Ren\textsuperscript{1,2,\textdagger}}}%
\thanks{\textsuperscript{1}The Chinese University of Hong Kong, Hong Kong SAR, China.}%
\thanks{\textsuperscript{2}Shenzhen Loop Area Institute, Shenzhen, China.}%
\thanks{\textsuperscript{3}Guangdong University of Technology, Guangzhou, China.}%
\thanks{\textsuperscript{*}Equal contribution. \textsuperscript{\textdagger}Corresponding author: Hongliang Ren.}}

\IEEEaftertitletext{%
\begin{center}%
\includegraphics[width=\textwidth,pagebox=cropbox,clip]{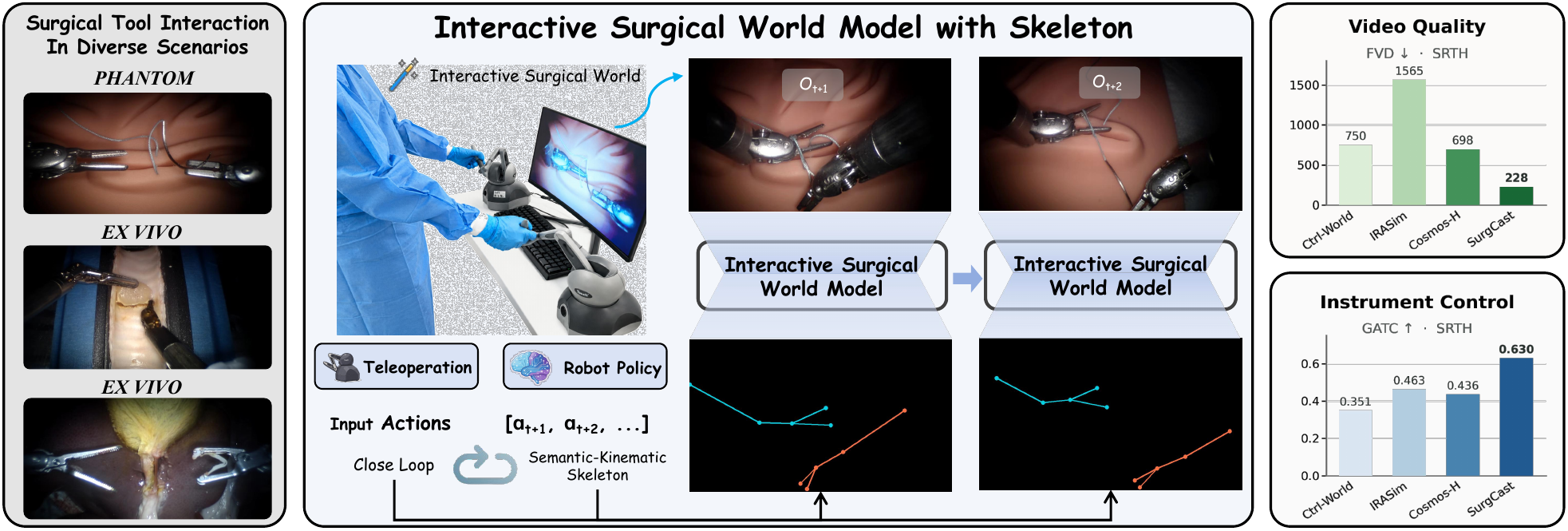}%
\refstepcounter{figure}\label{fig:teaser}%
\vspace{0.6\baselineskip}\\%
\parbox{0.98\textwidth}{\footnotesize\textbf{Fig.~\thefigure.} SurgCast forecasts future instrument skeletons from prospective robot actions and uses their geometry and derived kinematics to guide video generation. Read from surgical scenes (left), through commands and their structural and visual predictions (center), to quality and control summaries (right). The illustrated robot-policy loop motivates the framework; the evaluations use supplied action windows.}%
\end{center}%
}

\begin{document}
\raggedbottom

\maketitle
\thispagestyle{empty}
\pagestyle{empty}

\begin{abstract}
Surgical world models must turn prospective robot actions into visually
faithful, structurally consistent instrument motion. We introduce
\textbf{SurgCast}, which predicts future instrument skeletons from current
structure, robot state, and future actions, then uses the forecast to
control video generation. Dual-path injection combines skeleton geometry
with derived, instrument-specific motion descriptors through geometric
residuals and conditioning modulation. Both paths complement native
action conditioning and are retained in a four-step causal student.
On SutureBot and SRTH-Porcine-Cholei, the teacher and student improve
all seven reported quality and instrument-consistency metrics over their
respective direct-action baselines; teacher FVD decreases by 37.6\% and
67.3\%. Ablations show that future actions improve skeleton prediction
and that combining both injection paths performs best across all six
metrics under predicted skeletons. These results support using predicted
articulated structure and its motion semantics as a control interface
for surgical video prediction.
Project videos and examples are available at
https://surgcast.netlify.app/.

\end{abstract}

\section{Introduction}

A surgical robot's actions determine how instruments move within the
operative scene. An action-conditioned world model should make these
consequences visible before execution: different commands from the same
initial view should produce different, structurally consistent motions.
Visual prediction supports robot planning and interactive
simulation~\cite{deepvisualforesight2017,unipi2023,unisim2024}, while
action-conditioned models extend video generation to manipulation and
surgery~\cite{irasim2025,ctrlworld2026,cosmoshdreams2026}.
For surgery, realistic appearance must accompany instrument configurations
that follow the supplied commands.

Direct action conditioning leaves the generator to infer future
articulated geometry in image space. Our qualitative comparisons show
that recognizable scenes can coexist with instrument drift and
interaction misalignment (Fig.~\ref{fig:qualitative_comparison}).
Trajectory and kinematic controls expose structure to video
models~\cite{videocomposer2023,dragnuwa2023,motionctrl2024,saw2026,kvlr2026}.
Building on this principle, we study an explicit predictive interface:
forecast the instrument configuration from current structure and
prospective commands, then use that forecast to guide video generation.

Instrument skeletons connect these stages through keypoints with fixed
tool-specific roles. Their trajectories also yield descriptors of
image-plane translation, shaft-direction angular velocity, and projected
wrist and jaw articulation.
We call these instrument-meaningful motion descriptors
\emph{kinematic semantics}. Geometric maps locate the articulated tool,
while derived descriptors expose its motion directly to the generator's
conditioning. This raises two questions: do future commands improve
skeleton prediction, and does using the forecast's geometry and
kinematic semantics improve action-conditioned video generation?

\begin{figure*}[t]
  \centering
  \includegraphics[width=\textwidth]{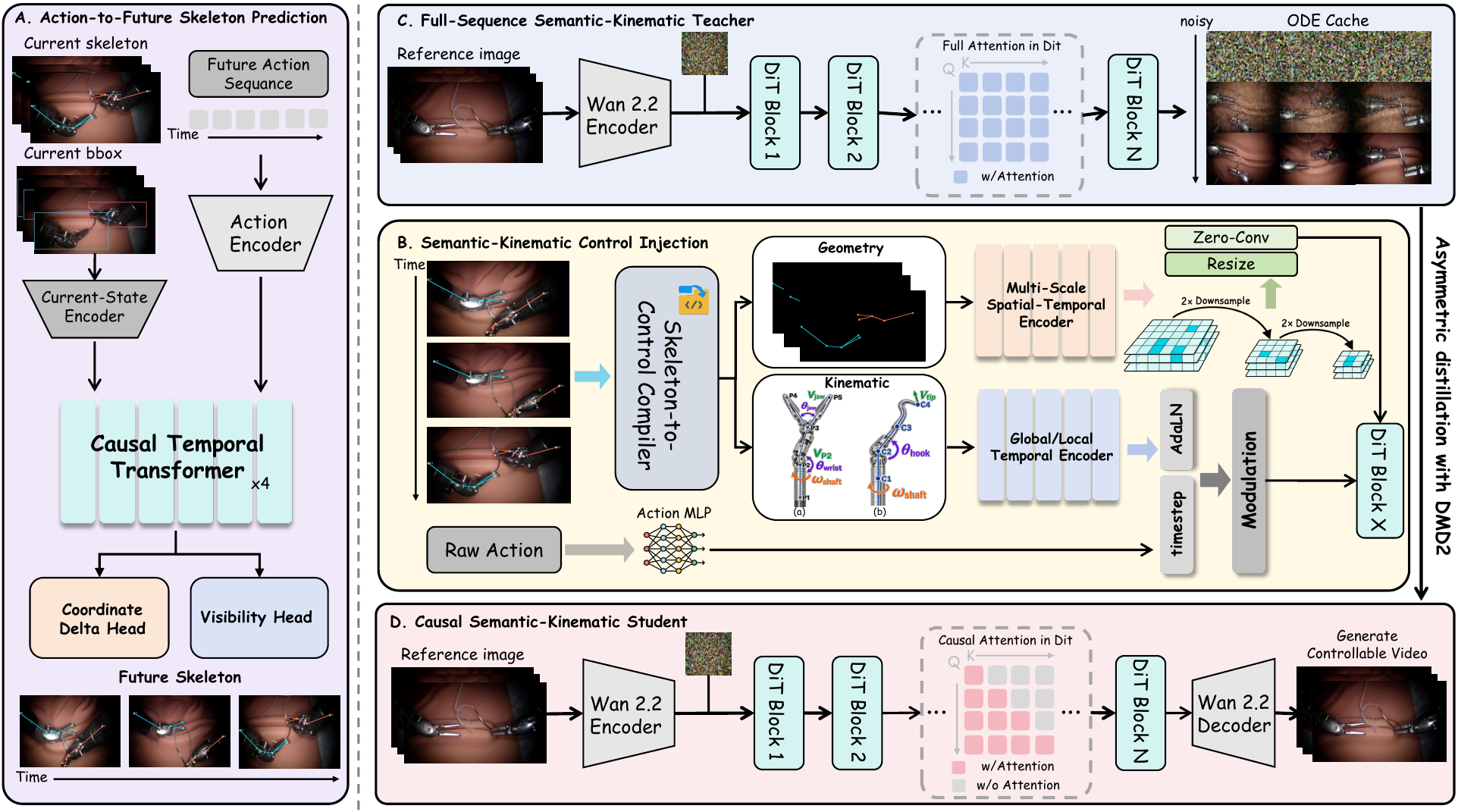}
  \caption{SurgCast's action-to-structure-to-video framework. Read
  (A)$\rightarrow$(B) for construction of the control interface:
  (A) forecasts future instrument skeletons from current structure and
  prospective actions; (B) converts the forecast into geometry maps for
  residual injection and derived kinematic descriptors for modulation.
  Both paths condition the teacher (C) and causal student (D), alongside
  native action inputs. The arrow from (C) to (D) denotes training-time
  distillation; teacher and student are alternative video generators.
  Kinematic quantities describe projected keypoint motion, including
  shaft-direction angular velocity in the image plane.}

  \label{fig:pipeline_overview}
\end{figure*}

\textbf{SurgCast} addresses both questions. An Action-to-Future-Skeleton
Predictor forecasts keypoints and visibility from current structure,
robot state, and future actions. Dual-Path Semantic--Kinematic Injection
converts the same forecast into geometry maps for spatial and temporal
residuals and motion descriptors for conditioning modulation.
The generator retains its native action inputs. This interface conditions
a full-sequence teacher and a four-step causal student trained through
distribution matching~\cite{dmd22024,causvid2025}.
Our contributions are:
\begin{itemize}
    \item An action-to-structure-to-video framework that uses predicted
    future instrument skeletons as an explicit control interface
    (Fig.~\ref{fig:pipeline_overview}; Table~\ref{tab:skeleton_prediction}).
    \item Dual-path injection of skeleton geometry and its derived
    motion semantics, with complementary benefits demonstrated under
    predicted skeletons (Table~\ref{tab:sk_ablation}).
    \item Retention of the interface in a four-step causal student,
    with improved video quality and instrument consistency over the
    corresponding direct-action baselines
    (Table~\ref{tab:overall_generation}; Fig.~\ref{fig:rollout_stability}).
\end{itemize}

\section{Related Work}

\subsection{Action-Conditioned Prediction for Robotics}
Visual foresight, generative policies, and interactive simulators connect
visual prediction to robot planning~\cite{deepvisualforesight2017,unipi2023,unisim2024}.
iVideoGPT and DINO-WM study interactive prediction and planning in visual
feature spaces~\cite{ivideogpt2024,dinowm2025}; IRASim, HMA, and Ctrl-World
model action-conditioned visual dynamics~\cite{irasim2025,hma2025,ctrlworld2026}.
In surgery, Open-H-Embodiment supplies paired video and
kinematics~\cite{openhembodiment2026}, and Cosmos-H-Dreams provides
action-conditioned generation and causal deployment~\cite{cosmoshdreams2026}.
Instrument pose and tracking supply articulated image-space
structure~\cite{mbapose2021,surgpose2025,tooltip2025}, while forecasting
methods predict motion or state from historical observations~\cite{davincinet2020,shi2022}.
SurgCast predicts future structure conditioned on prospective commands
and uses it as an intermediate control for video generation.

\subsection{Structured Control for Surgical Video Generation}
ControlNet and T2I-Adapter introduce spatial diffusion
conditioning~\cite{controlnet2023,t2iadapter2024}; VideoComposer,
DragNUWA, and MotionCtrl extend control to video
motion~\cite{videocomposer2023,dragnuwa2023,motionctrl2024}.
Surgical generators use action graphs, latent actions, object-aware
features, and tool-tip trajectories~\cite{visage2025,surgicalworldmodel2026,surgsora2025,saw2026}.
KVLR projects articulated kinematics using camera parameters into
image-aligned semantic, depth, orientation, velocity, and acceleration
controls and routes these modalities hierarchically~\cite{kvlr2026}.
Our control construction first forecasts future image-space skeletons
from current structure and commands. The forecast then supplies both
geometric maps and instrument-motion descriptors, injected as residuals
and conditioning modulation, respectively.

\subsection{Few-Step Causal Video Diffusion}
Progressive and distribution-matching distillation reduce sampling
cost~\cite{progressivedistill2022,dmd2024,dmd22024}.
Diffusion Forcing and CausVid study sequence and causal
generation~\cite{diffusionforcing2024,causvid2025}; Self Forcing,
Rolling Forcing, and Causal Forcing address self-generated histories
and extended rollouts~\cite{selfforcing2025,rollingforcing2025,causalforcing2026}.
SurgCast uses this training framework to retain its predicted structural
interface in a four-step causal generator.

\section{Method}

\subsection{Problem Formulation and Overview}
Given an initial image $I_0$, current instrument skeleton $S_0$, robot
state $r_0$, and prospective action sequence $a_{1:H}$, SurgCast generates
future observations $I_{1:H}$ whose
instrument configuration follows the supplied commands. Each instrument is
represented by five image-space keypoints with fixed structural roles and
visibility indicators. We denote the two-instrument structure by
$S_h=(P_h,V_h)$, with $P_h\in\mathbb{R}^{2\times5\times2}$.
The prediction and generation stages are
\begin{equation}
\left\{\begin{aligned}
\widehat S_{1:H} &= F_\phi(S_0,r_0,a_{1:H}),\\
I_{1:H} &\sim p_\theta(\,\cdot\mid I_0,a_{1:H},
                    \mathcal C(S_0,\widehat S_{1:H})).
\end{aligned}\right.
\label{eq:structure_interface}
\end{equation}
The operator $\mathcal C$ constructs two controls from the same forecast:
rasterized skeleton geometry and interpretable kinematic descriptors
derived from its keypoint trajectories (Fig.~\ref{fig:pipeline_overview}).
These descriptors express image-plane translation, shaft-direction
angular velocity, and projected wrist and jaw articulation.
The same structural interface conditions the full-sequence teacher
and causal student, both of which retain the backbone's original
action pathway and generate compressed video latents. The current skeleton is a supplied input; confidence, bounding boxes,
and tool identities accompany it as predictor inputs.

For long-horizon generation, each sequence contains 73 image frames:
one reference and 72 future frames. Temporal compression maps these to 19 video latents, with a
separate reference latent and 18 groups of four future frames. Structural
conditions are temporally pooled onto this same grid. We use $h$ for
image/skeleton time, $j$ for latent time, and $\sigma$ for diffusion noise
level throughout.

\subsection{Action-to-Future-Skeleton Prediction}
\paragraph{Anchored structural forecasting}
The predictor uses the current configuration as a common reference for
future motion. We normalize keypoint coordinates to $[-1,1]$. A bounding-box-conditioned
completion module first estimates missing current keypoints from visible
structure, confidence, robot state, and tool identity. Observed keypoints
are preserved; completed points provide anchors for occluded locations
with valid bounding boxes. Let $\widetilde P_0$ denote the resulting current
configuration. The predictor embeds this configuration and robot state into
a context vector $c_0$. At step $h$, the input token is
$u_h=E_a(a_h)+c_0+e_h$, where $e_h$ is a learned forecast-horizon embedding.
A causally masked transformer computes
$q_{1:H}=\operatorname{Tr}_{\phi}(u_{1:H})$. Its output heads predict
coordinates and visibility:
\begin{equation}
\left\{\begin{aligned}
\widehat P_h &= \widetilde P_0+D_\phi(q_h),\\
\widehat V_h &= \operatorname{sigmoid}(W_vq_h).
\end{aligned}\right.
\label{eq:predictor}
\end{equation}
A causal attention mask makes $q_h$ depend on commands up to step $h$, with a padding
mask for variable-length sequences. Coordinates are predicted as offsets
from the completed current configuration instead of accumulated
frame-to-frame displacements. The implemented transformer has four layers,
width 256, and eight attention heads. Per instrument, the predictor
represents commands as three relative translation values, a six-dimensional
relative rotation, and one absolute jaw value, normalized using training
statistics. The video backbone separately constructs the corresponding
20-dimensional two-arm action representation and pads it to 44 dimensions.

\paragraph{Visibility-aware learning}
The joint prediction objective is
\begin{equation}
\mathcal L_{\rm pred}
=\mathcal L_{\rm coord}+0.2\mathcal L_{\rm disp}
 +0.1\mathcal L_{\rm vis}+0.5\mathcal L_{\rm comp}.
\label{eq:predictor_loss}
\end{equation}
Coordinate ($\mathcal L_{\rm coord}$) and current-keypoint completion
($\mathcal L_{\rm comp}$) use Smooth-$L_1$ with threshold 0.02. $\mathcal L_{\rm disp}$ applies the same penalty
to consecutive-coordinate differences, using pairs with valid targets.
Coordinate supervision is confidence-weighted and normalized within each
instrument before combining instrument losses; displacement weights use the
minimum confidence of each pair. Visibility loss $\mathcal L_{\rm vis}$ uses binary cross-entropy on
known labels. Completion supervision targets artificially masked current
keypoints. To handle interchangeable jaw-tip labels, the coordinate loss
selects the lower-error identity or jaw-tip-swapped assignment for each
instrument over the entire prediction horizon, and applies that assignment
consistently to the corresponding targets. The trained predictor supplies
structural conditions to video generation.

\subsection{Dual-Path Semantic--Kinematic Injection}
The two paths give the generator complementary access to the same
predicted skeleton. Geometric maps preserve the spatial support of
keypoints and edges and encode their temporal changes as localized
residuals. Derived kinematic descriptors expose instrument motion in a
compact form for conditioning modulation. This division lets structure
guide video tokens while instrument-meaningful motion quantities guide
the conditioning. Global and local descriptors are groups within the
kinematic path. Both paths complement the native 44-dimensional action
interface.

\paragraph{Multi-scale geometric residuals}
This path preserves the spatial support and temporal evolution of the
skeleton. For each instrument, we rasterize five keypoints and the four edges
$(1,2)$, $(2,3)$, $(3,4)$, and $(3,5)$ as Gaussian point and segment-distance
maps. Corresponding validity channels distinguish absent observations from
empty spatial support. The two instruments yield 18 geometric channels and
18 validity channels. After temporal alignment, these form
$G_j\in\mathbb{R}^{36\times36\times64}$ on the latent spatial grid.

Separate convolutional stems encode node, edge, and validity maps. Their
features are fused into a three-level, width-512 spatial pyramid
$F_j^{(\ell)}$. At each level, a temporal transformer encodes consecutive

feature differences $\Delta F_j^{(\ell)}=F_j^{(\ell)}-F_{j-1}^{(\ell)}$,
with zero difference at the first step. At selected backbone block $b$, spatial and temporal residuals jointly
update video tokens:
\begin{equation}
\left\{\begin{aligned}
R_{b,j}^{s}&=Z_b^{s}\!\left(U_b(F_j^{(\ell_b)})\right),\\
R_{b,j}^{t}&=Z_b^{t}\!\left(
 U_b(\operatorname{Tr}_g(\Delta F^{(\ell_b)})_j)\right),\\
\widetilde X_{b,j}&=X_{b,j}+R_{b,j}^{s}+R_{b,j}^{t}.
\end{aligned}\right.
\label{eq:geometry_injection}
\end{equation}
$U_b$ resizes features to the backbone token grid and $Z_b^s,Z_b^t$ are
zero-initialized projections to its hidden dimension. Validity gates mask
unavailable controls, and the reference latent receives no generated-frame
residual. In the 28-block backbone, injection occurs before blocks
1, 8, 15, 22, and 28, using pyramid levels 1, 1, 2, 3, and 3, respectively.

\paragraph{Global--local kinematic modulation}
This path gives projected motion explicit instrument meanings. We
derive nine descriptors per instrument from image-plane keypoints:
the two-dimensional velocity of $P_2$, the angular velocity of the
projected shaft direction, projected wrist angle and angular velocity,
projected jaw opening angle and angular velocity, and the two-dimensional
velocity of the jaw midpoint. Shaft angular velocity measures the
change of the $P_1P_2$ direction in the image plane. Velocities divide
coordinate or wrapped angular differences by elapsed time. Each
descriptor has a validity indicator; valid values are temporally pooled
and standardized using training-set statistics, and invalid entries
are set to zero.

The global descriptor group contains translation and shaft-direction change
(six values across two instruments); the local group contains wrist and jaw
articulation and jaw-midpoint motion (twelve values). Appending validity
indicators gives 12- and 24-dimensional inputs, respectively. Separate
width-256 temporal encoders process embedded global and local descriptor
sequences $\kappa^g$ and $\kappa^l$. Their outputs at latent step $j$ are
concatenated as $k_j$. Let $e_\sigma,m_\sigma$ denote noise-level terms and
$e_j^a,m_j^a$ the original action terms. Kinematics enters the backbone as
\begin{equation}
\left\{\begin{aligned}
e_j &= \operatorname{Norm}
       \big(e_\sigma+e_j^a+Z_e(k_j)\big),\\
m_j &= m_\sigma+m_j^a+Z_m(k_j).
\end{aligned}\right.
\label{eq:kinematic_injection}
\end{equation}
Here, $e_j$ is the backbone conditioning embedding and $m_j$ is its adaptive
layer-normalization modulation input; superscript $a$ denotes the original
action pathway. The zero-initialized projections $Z_e$ and $Z_m$ output 2048
and $3\times2048$ channels. All control temporal encoders use two layers and eight heads.
Zero-initialized output projections preserve the pretrained backbone's
function at the start of control training.

\setlength{\dbltextfloatsep}{8pt plus 2pt minus 2pt}
\begin{table*}[!t]
\centering
\begingroup
\caption{Generation quality on SutureBot and SRTH-Porcine-Cholei,
with teachers and four-step students grouped separately. SurgCast
(Predicted) is the main method and uses forecast future skeletons;
SurgCast (Oracle) supplies ground-truth future skeletons as a reference
condition. These labels specify the skeleton source, while the teacher
and student groups specify the video-generation mode.
The causal student uses EMA-600. Green (bold) and yellow mark the best
and second-best results per group and dataset.
Cosmos-Predict2.5-2B results are from~\cite{cosmospredict252025}.}
\label{tab:overall_generation}
\fontsize{5.8}{7.0}\selectfont
\setlength{\tabcolsep}{1.55pt}
\renewcommand{\arraystretch}{1.14}

\begin{tabular*}{\textwidth}{@{\extracolsep{\fill}} l*{14}{c} @{}}
\toprule
\raisebox{-0.8em}{\textbf{Method}}
& \multicolumn{7}{c}{\cellcolor{suturebotheader}\textbf{SutureBot}}
& \multicolumn{7}{c}{\cellcolor{srthheader}\textbf{SRTH-Porcine-Cholei}} \\
\cmidrule(lr){2-8}\cmidrule(lr){9-15}
& \textbf{PSNR}$\uparrow$ & \textbf{SSIM}$\uparrow$ & \textbf{LPIPS}$\downarrow$
& \textbf{FID}$\downarrow$ & \textbf{FVD}$\downarrow$ & \textbf{GATC}$\uparrow$ & \textbf{TCD}$\downarrow$
& \textbf{PSNR}$\uparrow$ & \textbf{SSIM}$\uparrow$ & \textbf{LPIPS}$\downarrow$
& \textbf{FID}$\downarrow$ & \textbf{FVD}$\downarrow$ & \textbf{GATC}$\uparrow$ & \textbf{TCD}$\downarrow$ \\
\midrule

\rowcolor{groupgray}
\multicolumn{15}{c}{\textbf{\textit{Teachers}}} \\
Cosmos-Predict2.5-2B~\cite{cosmospredict252025}
& 8.1037 & 0.2249 & 0.7951 & 394.0091 & 6815.0560 & 0.0036 & 293.72
& 10.4087 & 0.1888 & 0.7492 & 458.5920 & 9802.3112 & 0.1340 & 293.7209 \\
Ctrl-World~\cite{ctrlworld2026}
& 21.5191 & 0.7856 & 0.2321 & 49.6254 & 248.3254 & 0.3656 & 71.95
& 19.3131 & 0.6584 & 0.3222 & 88.1025 & 750.3728 & 0.3507 & 70.3725 \\
IRASim~\cite{irasim2025}
& 17.9001 & 0.5556 & 0.4226 & 135.8555 & 1226.6812 & 0.3551 & 68.04
& 16.6724 & 0.3632 & 0.4786 & 218.6194 & 1564.5107 & 0.4632 & 81.2678 \\
Cosmos-H-Surgical-Simulator~\cite{cosmoshdreams2026}
& 22.1164 & 0.8041 & 0.1719
& 31.7472 & 172.4712
& 0.6522 & 47.59
& 18.6471 & 0.6387 & 0.3027 & 92.6880 & 698.0070 & 0.4363 & 63.5268 \\
SurgCast (Predicted)
& \cellcolor{bestyellow}24.4687 & \cellcolor{bestyellow}0.8458 & \cellcolor{bestyellow}0.1073
& \cellcolor{bestyellow}27.0716 & \cellcolor{bestyellow}107.5858
& \cellcolor{bestyellow}0.7902 & \cellcolor{bestyellow}40.37
& \cellcolor{bestyellow}21.3761 & \cellcolor{bestyellow}0.7530 & \cellcolor{bestyellow}0.1523 & \cellcolor{bestyellow}37.0874 & \cellcolor{bestyellow}228.1920 & \cellcolor{bestyellow}0.6298 & \cellcolor{bestyellow}49.7813 \\
SurgCast (Oracle)
& \cellcolor{oursgreen}\textbf{25.0237} & \cellcolor{oursgreen}\textbf{0.8534} & \cellcolor{oursgreen}\textbf{0.1015} & \cellcolor{oursgreen}\textbf{26.0625}
& \cellcolor{oursgreen}\textbf{104.4354} & \cellcolor{oursgreen}\textbf{0.8109} & \cellcolor{oursgreen}\textbf{38.75}
& \cellcolor{oursgreen}\textbf{21.4431} & \cellcolor{oursgreen}\textbf{0.7546} & \cellcolor{oursgreen}\textbf{0.1511} & \cellcolor{oursgreen}\textbf{36.7987}
& \cellcolor{oursgreen}\textbf{222.0160} & \cellcolor{oursgreen}\textbf{0.6318} & \cellcolor{oursgreen}\textbf{48.1370} \\

\rowcolor{groupgray}
\multicolumn{15}{c}{\textbf{\textit{Students}}} \\
Cosmos-H-Surgical-Simulator (Original Student)
& \cellcolor{bestyellow}22.1901 & \cellcolor{bestyellow}0.7908 & \cellcolor{bestyellow}0.1943 & \cellcolor{bestyellow}59.0538 & \cellcolor{bestyellow}212.5411 & \cellcolor{bestyellow}0.5151 & \cellcolor{bestyellow}60.36
& \cellcolor{bestyellow}17.5261 & \cellcolor{bestyellow}0.5932 & \cellcolor{bestyellow}0.4224 & \cellcolor{bestyellow}92.3297 & \cellcolor{bestyellow}700.6836 & \cellcolor{bestyellow}0.3236 & \cellcolor{bestyellow}82.0362 \\
SurgCast (Causal Student)
& \cellcolor{oursgreen}\textbf{23.2639} & \cellcolor{oursgreen}\textbf{0.8114} & \cellcolor{oursgreen}\textbf{0.1504}
& \cellcolor{oursgreen}\textbf{50.1808} & \cellcolor{oursgreen}\textbf{191.5575} & \cellcolor{oursgreen}\textbf{0.6278} & \cellcolor{oursgreen}\textbf{47.50}
& \cellcolor{oursgreen}\textbf{20.2579} & \cellcolor{oursgreen}\textbf{0.7134} & \cellcolor{oursgreen}\textbf{0.1753}
& \cellcolor{oursgreen}\textbf{49.7365} & \cellcolor{oursgreen}\textbf{315.0477} & \cellcolor{oursgreen}\textbf{0.5411} & \cellcolor{oursgreen}\textbf{53.7209} \\
\bottomrule
\end{tabular*}
\endgroup
\end{table*}

\begin{figure*}[!t]
  \centering
  \includegraphics[width=\textwidth,pagebox=cropbox]{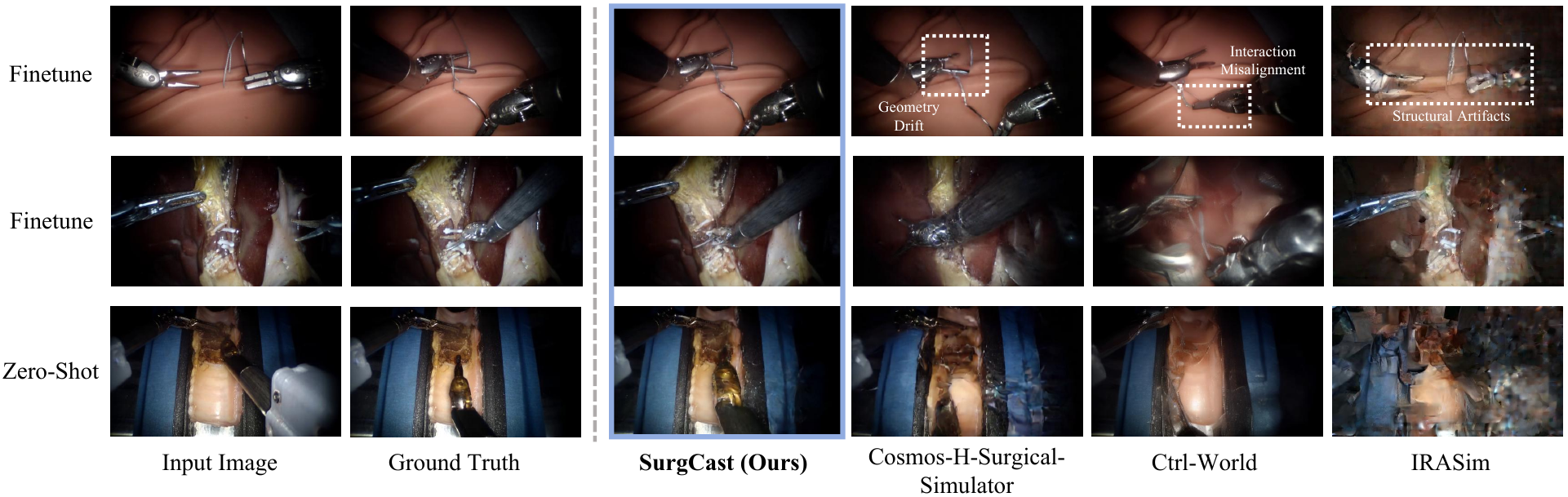}
  \caption{Teacher outputs on SutureBot (top), SRTH-Porcine-Cholei
  (middle), and Cao\_cautery\_combin with zero-shot video-generator
  transfer (bottom). The figure label
  'SurgCast (Ours)' denotes SurgCast (Predicted), conditioned on predicted
  future skeletons. White dashed annotations
  identify geometry drift, interaction misalignment, and structural
  artifacts in baseline outputs; compare the annotated instrument regions
  with SurgCast under the corresponding scene and action sequence.}
  \label{fig:qualitative_comparison}
\end{figure*}

\subsection{Teacher Training and Causal Distillation}
\paragraph{Structurally conditioned teacher}
Let $z$ denote clean video latents and let $c$ collect the reference image,
actions, and structural conditions. For Gaussian noise
$\epsilon\sim\mathcal N(0,\mathbf I)$, the interpolated latent is
$z_\sigma=(1-\sigma)z+\sigma\epsilon$. We train the full-sequence teacher
with the backbone's rectified-flow objective:
\begin{equation}
\mathcal L_{\rm RF}=\mathbb E\!\left[
 w(\sigma)\left\|v_\theta(z_\sigma,\sigma;c)
                   -(\epsilon-z)\right\|_{\rm mean}^{2}\right].
\label{eq:teacher_rf}
\end{equation}
The notation $\|\cdot\|_{\rm mean}^{2}$ denotes mean squared error over
tensor elements, and $w$ is the backbone's noise-level weighting.
The reference frame is observed. Predicted conditioning uses a frozen
skeleton predictor. The Oracle reference is trained separately with
annotated future structure from the same initialization. The injection
ablation uses predicted skeletons and retains native action conditioning.

\paragraph{Causal student with the same structural interface}
To carry the structural interface into autoregressive video generation,
the student retains both injection paths and factorizes the $J=18$
future video latents as
\begin{equation}
p_\psi(z_{1:J}\mid z_0,c)
=\prod_{j=1}^{J}p_\psi(z_j\mid z_0,z_{1:j-1},c).
\label{eq:causal_factorization}
\end{equation}
Here, $\psi$ denotes student parameters and $z_0$ the reference latent.
Cached video attention accesses only preceding generated latents.
Structural encoders process the supplied prospective control window $c$;
the student retrieves features aligned with each latent. Causality here
concerns generated video history, with the prospective controls supplied
for the prediction window.

We first warm-start the student using teacher ODE trajectories, regressing
its clean-latent estimates to the trajectory endpoint on future frames.
We then use asymmetric distribution matching, following
DMD2~\cite{dmd22024} and bidirectional-to-causal
distillation~\cite{causvid2025}, with self-generated rollout histories as in
Self Forcing~\cite{selfforcing2025}. Let $y$ be a student-generated latent
sequence. In the distillation parameterization, it is re-noised as
$x_\tau=y\cos\tau+\eta\sin\tau$, where $\eta$ is Gaussian noise.
A frozen bidirectional teacher $D_T$ and a trainable bidirectional
student-distribution denoiser $D_f$ predict clean latents under the same
condition $c$, yielding $d_T=D_T(x_\tau,\tau;c)$ and
$d_f=D_f(x_\tau,\tau;c)$. Define $g=(d_f-d_T)/s$, with
$s=\max(\operatorname{mean}|y-d_T|,10^{-5})$. The generator surrogate is
\begin{equation}
\mathcal L_{\rm DMD}
=\mathbb E\left[\|y-\operatorname{sg}(y-g)\|_{\rm mean}^{2}\right].
\label{eq:dmd}
\end{equation}
Here $\operatorname{sg}$ stops gradients. The auxiliary denoiser
reconstructs detached student samples with $\sin^{-2}\tau$-weighted
squared error. We perform five auxiliary-denoiser updates per generator
update and use four denoising steps at inference.

\section{Experiments}

\subsection{Experimental Setup}
We evaluate the action-to-structure-to-video chain on SutureBot and
SRTH-Porcine-Cholei, with Cao\_cautery\_combin for zero-shot
video-generator transfer. Zero-shot denotes a video generator evaluated
without fine-tuning on Cao; the three-domain predictor benchmark below
uses a separate training protocol.
SurgCast (Predicted), our main method, uses forecast future skeletons;
Oracle supplies ground-truth future skeletons as a reference. The
injection ablation also uses predicted skeletons and retains native
action conditioning.

SutureBot has 1161/145/146 disjoint training/validation/test episodes
and 47,710 complete H73 training windows. Its matched video panel
contains 28 eligible validation episodes, each yielding one 72-frame
forecast; SRTH teacher comparisons use 95 test episodes and 24 future
frames. Both use seed 0, $512\times288$ resolution, and stride five
(approximately 6\,Hz for 30\,Hz SutureBot recordings).

SutureBot structural teachers share the Cosmos-H DiT initialization and
Wan tokenizer, with zero-initialized control outputs. Training uses four
H100 GPUs, effective batch 64, and up to 1600 AdamW updates
(learning rate $4\times10^{-5}$, weight decay 0.1).
Teachers use 35 sampling steps with guidance 0. Ctrl-World and IRASim
use 50 steps with guidance 1 and native action/temporal interfaces,
adapted on the same SutureBot training partition.
SutureBot current-frame keypoints, visibility, confidence, and boxes come
from offline framewise SurgPose detections; robot state comes from the
current observation and tool identities from episode metadata. A frozen predictor forecasts 72 steps jointly
from these current inputs and future actions.

\begin{figure*}[t]
  \centering
  \includegraphics[width=\textwidth,pagebox=cropbox]{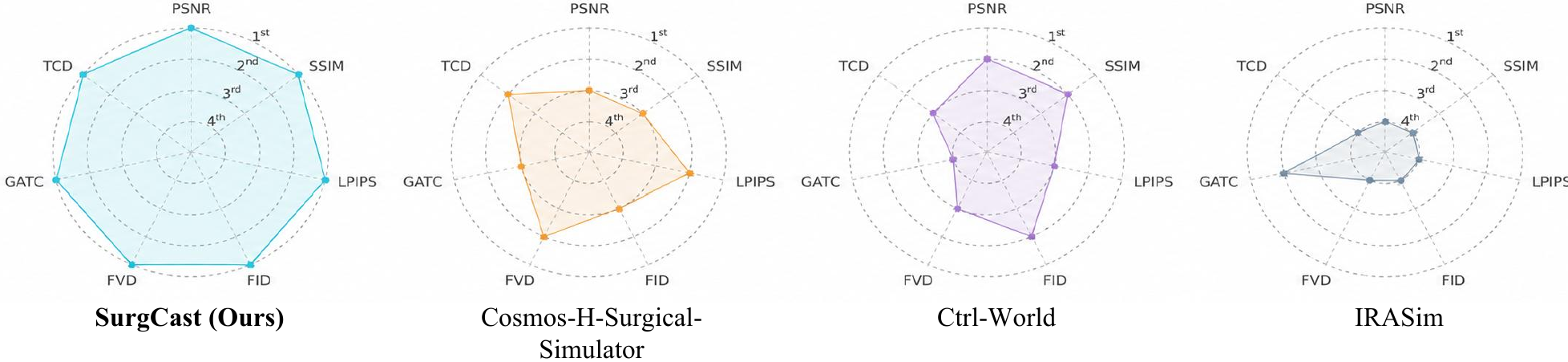}
  \caption{Video-generator transfer to Cao\_cautery\_combin without
  video-model fine-tuning: ranks across seven quality and instrument-
  consistency metrics. 'SurgCast (Ours)' labels the SurgCast teacher;
  the outer ring denotes rank 1. The radial
  positions encode ranking within each metric, not metric magnitudes.}
  \label{fig:cao_cautery_combin_zeroshot_radar}
\end{figure*}

PSNR, SSIM, and LPIPS assess frame fidelity; FID and FVD assess image-
and video-level quality. FVD uses the first 16 future frames in the
SutureBot and SRTH comparisons. GT-Anchored Tool Consistency (GATC) compares
grayscale appearance in ground-truth tool regions using zero-mean
normalized cross-correlation, a gradient-presence penalty, and local
alignment within $\pm3$ pixels and $\pm1$ frame. Tool Centroid Distance
(TCD) matches Medical-SAM3 tool-mask centroids by Hungarian assignment;
a missing tool incurs half the image diagonal. Both use medians over
valid future frames and exclude the observed first frame. For rollout
evaluation, C1--C6 partition the
72 future frames into six consecutive 12-frame segments. FDS-L1 and
FDS-SSIM report segment-averaged framewise L1 error and SSIM against
the ground-truth video, with RGB values normalized to $[-1,1]$; GATC and
TCD are also reported by segment.

Table~\ref{tab:skeleton_prediction} macro-averages H12 prediction over
SutureBot, SRTH-Porcine-Cholei, and Cao\_cautery\_combin, each with
1161/145/146 training/validation/test windows from disjoint episodes.
It evaluates a jointly trained model without adaptation and separate
domain-adapted models over 12 future steps at stride five. ADE and FDE
measure mean and final-step error in $960\times540$ coordinates;
PCK@20px is the fraction within 20 pixels. The 72-step video controls
use a separately trained predictor. Throughput and peak GPU memory
are measured over 10 four-step, 73-frame rollouts.

\subsection{Generation Quality and Instrument Consistency}
We first ask whether the predicted structural interface improves
action-conditioned video generation. Table~\ref{tab:overall_generation}
compares SurgCast (Predicted), our main method, with the direct-action
teacher. It improves all seven reported metrics on both
datasets. FVD decreases by 37.6\% on SutureBot and 67.3\% on
SRTH-Porcine-Cholei, alongside improved GATC and TCD. The paired
results show improved video quality together with closer instrument
appearance and position under the supplied actions.

The separately trained Oracle reference uses ground-truth future
structure. Predicted and Oracle yield FVD values
of 107.59 and 104.44 on SutureBot, and 228.19 and 222.02 on
SRTH-Porcine-Cholei. Relative to the gains over direct-action
conditioning, the modest differences show that predicted structure
retains much of the generation benefit observed with annotated structure.

The four-step causal student tests whether the interface carries
through distillation. It improves all seven metrics over the
original student, with FVD decreasing from 212.54 to 191.56
on SutureBot and from 700.68 to 315.05 on SRTH-Porcine-Cholei.
This comparison extends the full-sequence result to causal
generation under the same four-step student sampling budget.

Figure~\ref{fig:qualitative_comparison} illustrates how the predicted-skeleton
teacher preserves instrument structure against baseline geometry drift,
interaction misalignment, and structural artifacts. These examples
complement the Predicted results in Table~\ref{tab:overall_generation}.

\subsection{Ablation Studies}
The ablations examine both links in the structural interface:
how geometric and kinematic injection use the forecast, and what
prospective commands contribute to skeleton prediction.
\paragraph{Dual-path semantic--kinematic injection}
Table~\ref{tab:sk_ablation} tests the two injection paths under predicted
skeletons in the SutureBot teacher. The action-only control uses the
same training recipe; all variants retain native action conditioning.
Either structural path improves all six metrics over action-only.
Geometry yields lower LPIPS and FID, while kinematics yields higher
PSNR, SSIM, and GATC and lower TCD. Combining both paths performs
best on every metric, reaching PSNR of 24.4687, LPIPS of 0.1073,
and TCD of 40.37. The combined row is SurgCast (Predicted) in
Table~\ref{tab:overall_generation}. These results support complementary
use of geometric residuals and derived motion semantics in the main
predicted-skeleton generation pathway.

\begin{table}[!tb]

\centering
\caption{Predicted-skeleton ablation in the SutureBot teacher.
Enabled paths use forecast skeletons; all variants, including action-only,
use the same training recipe and retain native action conditioning.
Yellow marks the best single-path result; green marks the combined model.}
\label{tab:sk_ablation}
\setlength{\tabcolsep}{2.0pt}
\renewcommand{\arraystretch}{1.12}
\resizebox{\columnwidth}{!}{%
\begin{tabular}{ccccccccc}
\toprule
\textbf{Method} & \textbf{Geo.} & \textbf{Kin.}
& \textbf{PSNR}$\uparrow$ & \textbf{SSIM}$\uparrow$
& \textbf{LPIPS}$\downarrow$ & \textbf{FID}$\downarrow$
& \textbf{GATC}$\uparrow$ & \textbf{TCD}$\downarrow$ \\
\midrule
Action only
& --- & --- & 22.1164 & 0.8041 & 0.1719 & 31.7472 & 0.6522 & 47.59 \\
Geometry only
& $\checkmark$ & ---
& 23.0351 & 0.8337
& \cellcolor{bestyellow}0.1212 & \cellcolor{bestyellow}27.8214
& 0.7815 & 41.7564 \\
Kinematics only
& --- & $\checkmark$
& \cellcolor{bestyellow}23.5359 & \cellcolor{bestyellow}0.8368
& 0.1370 & 28.8493
& \cellcolor{bestyellow}0.7869 & \cellcolor{bestyellow}41.3922 \\
\midrule
\begin{tabular}[c]{@{}c@{}}Geometry +\\Kinematics\end{tabular}
& $\checkmark$ & $\checkmark$
& \cellcolor{oursgreen}\textbf{24.4687}
& \cellcolor{oursgreen}\textbf{0.8458}
& \cellcolor{oursgreen}\textbf{0.1073}
& \cellcolor{oursgreen}\textbf{27.0716}
& \cellcolor{oursgreen}\textbf{0.7902}
& \cellcolor{oursgreen}\textbf{40.37} \\
\bottomrule
\end{tabular}%
}
\end{table}

\begin{figure}[!tb]
  \centering
  \includegraphics[width=\columnwidth,pagebox=cropbox]{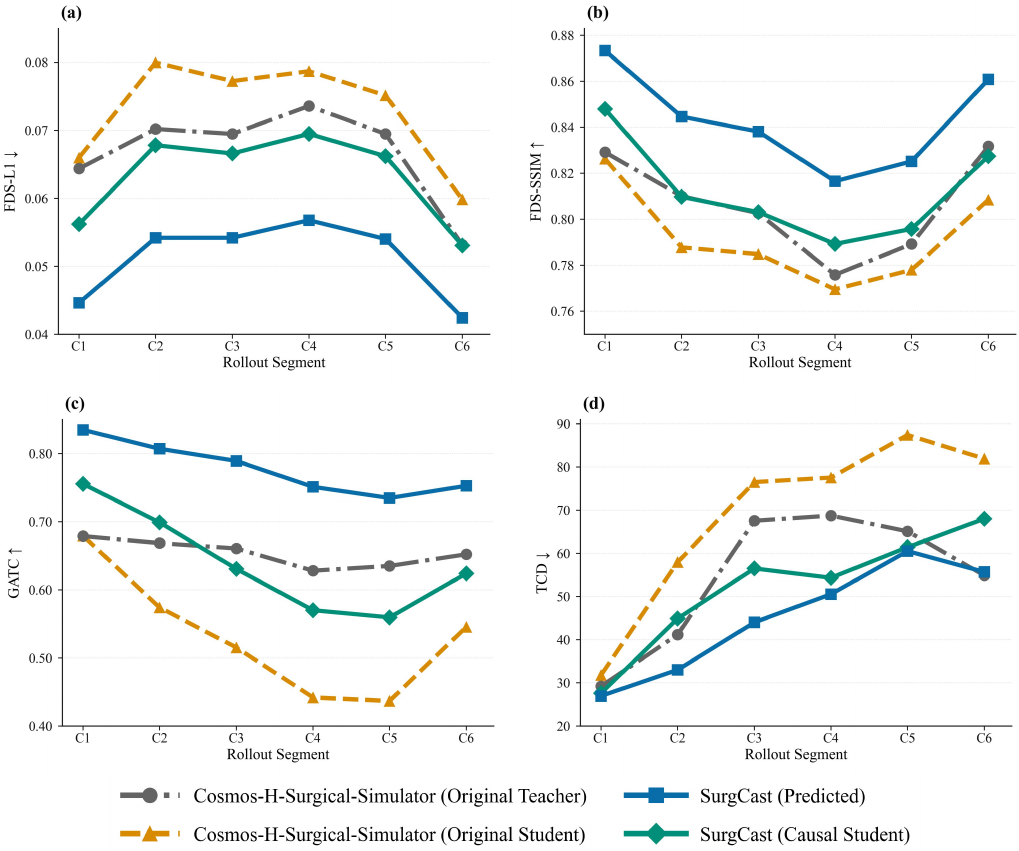}
  \caption{Segment-wise rollout on 28 SutureBot validation episodes.
  C1--C6 cover successive 12-frame portions of the 72-frame prediction.
  Blue squares denote SurgCast (Predicted), conditioned on predicted future
  skeletons; student curves compare four-step causal generators. Panels (a,b) report frame fidelity;
  (c,d) report tool-region consistency and centroid distance.}
  \label{fig:rollout_stability}
\end{figure}

\paragraph{Action-to-future-skeleton prediction}
Table~\ref{tab:skeleton_prediction} tests what future actions add to
current structure and robot state. For the joint three-domain predictor,
actions reduce ADE from 29.30 to 15.94 and FDE@H12 from 41.33 to
20.03, while PCK@20px rises from 57.65\% to 75.08\%. All four
metrics also improve over persistence.

Domain-adapted H12 predictors improve ADE and PCK in every interval.
At steps 9--12, actions reduce ADE from 38.82 to 19.07 and increase
PCK from 48.06\% to 68.45\% over prediction without future actions.

The 72-step control predictor is separately evaluated on 732 windows
from 28 SutureBot validation episodes using $960\times540$ pose
pseudo-labels. Its valid-point ADE is 30.76 pixels, versus 96.07 for
a separately trained no-action model; over steps 37--72, the errors
are 32.38 and 119.59 pixels. This supports action-conditioned
forecasting at the full video horizon.

\begin{table}[!tb]

\centering
\caption{H12 skeleton prediction: three-domain macro-average in
$960\times540$ coordinates. Top: jointly trained predictor without further
adaptation. Bottom: domain-adapted predictors over steps 1--4, 5--8,
and 9--12 (stride 5). Inputs cumulatively add $r_0$ and $a_{1:H}$ to
$S_0$. Yellow marks the best baseline; green marks the action-conditioned predictor.}
\label{tab:skeleton_prediction}
\setlength{\tabcolsep}{2.0pt}
\renewcommand{\arraystretch}{1.12}

\resizebox{\columnwidth}{!}{%
\begin{tabular}{cccccc}
\toprule
\rowcolor{groupgray}
\multicolumn{6}{c}{\textbf{\textit{Overall Performance}}} \\
\textbf{Method} & \textbf{Input}
& \textbf{ADE}$\downarrow$ & \textbf{FDE@H12}$\downarrow$
& \textbf{PCK@20px}$\uparrow$ & \textbf{Vis. Acc.}$\uparrow$ \\
\midrule
Persistence
& $S_0$ & 55.64 & 80.83 & 49.04\% & 92.96\% \\
\begin{tabular}[c]{@{}c@{}}Predictor w/o\\future action\end{tabular}
& $+r_0$
& \cellcolor{bestyellow}29.30 & \cellcolor{bestyellow}41.33
& \cellcolor{bestyellow}57.65\% & \cellcolor{bestyellow}97.33\% \\
\begin{tabular}[c]{@{}c@{}}Action-to-Skeleton\\Predictor\end{tabular}
& $+a_{1:H}$
& \cellcolor{oursgreen}\textbf{15.94} & \cellcolor{oursgreen}\textbf{20.03}
& \cellcolor{oursgreen}\textbf{75.08\%} & \cellcolor{oursgreen}\textbf{97.89\%} \\
\bottomrule
\end{tabular}%
}

\vspace{0.7ex}

\resizebox{\columnwidth}{!}{%
\begin{tabular}{ccccccc}
\toprule
\rowcolor{groupgray}
\multicolumn{7}{c}{\textbf{\textit{Performance by Horizon}}} \\
\textbf{Method}
& \multicolumn{3}{c}{\textbf{ADE}$\downarrow$}
& \multicolumn{3}{c}{\textbf{PCK@20px}$\uparrow$} \\
\cmidrule(lr){2-4}\cmidrule(lr){5-7}
& Early & Mid & Late & Early & Mid & Late \\
\midrule
Persistence
& 30.43 & 59.95 & 76.71 & 62.98\% & 44.66\% & 39.36\% \\
\begin{tabular}[c]{@{}c@{}}Predictor w/o\\future action\end{tabular}
& \cellcolor{bestyellow}18.34 & \cellcolor{bestyellow}30.87
& \cellcolor{bestyellow}38.82 & \cellcolor{bestyellow}71.18\%
& \cellcolor{bestyellow}53.58\% & \cellcolor{bestyellow}48.06\% \\
\begin{tabular}[c]{@{}c@{}}Action-to-Skeleton\\Predictor\end{tabular}
& \cellcolor{oursgreen}\textbf{12.42} & \cellcolor{oursgreen}\textbf{16.38}
& \cellcolor{oursgreen}\textbf{19.07} & \cellcolor{oursgreen}\textbf{82.77\%}
& \cellcolor{oursgreen}\textbf{73.91\%} & \cellcolor{oursgreen}\textbf{68.45\%} \\
\bottomrule
\end{tabular}%
}
\end{table}

\subsection{Temporal Consistency and Video-Generator Transfer}
\paragraph{Temporal rollout}
Figure~\ref{fig:rollout_stability} examines structural conditioning
across six SutureBot rollout segments. SurgCast (Predicted) has
lower FDS-L1 and higher FDS-SSIM and GATC
than the direct-action teacher in C1--C6; its TCD is lower in the
first five segments and slightly higher at C6.

The causal student improves all four reported metrics over the
original student in every segment. These within-segment gains show that
the structural interface remains beneficial across the evaluated
rollout, while the nonmonotonic curves reflect variation between
segments.

\paragraph{Zero-shot video-generator transfer}
On Cao\_cautery\_combin, without video-generator fine-tuning,
Fig.~\ref{fig:cao_cautery_combin_zeroshot_radar} ranks the
SurgCast teacher against Cosmos-H-Surgical-Simulator,
Ctrl-World, and IRASim. SurgCast ranks first on PSNR, SSIM,
LPIPS, FID, FVD, GATC, and TCD. The bottom row of
Fig.~\ref{fig:qualitative_comparison} shows the corresponding
visual comparison. Together, the ranks and example extend
the evaluated generation benefit to this video-generator transfer setting.

\subsection{Operator-Driven Interaction}
\begin{figure}[!tb]

  \centering
  \vspace*{3pt}
  \includegraphics[width=\columnwidth,pagebox=cropbox,clip]{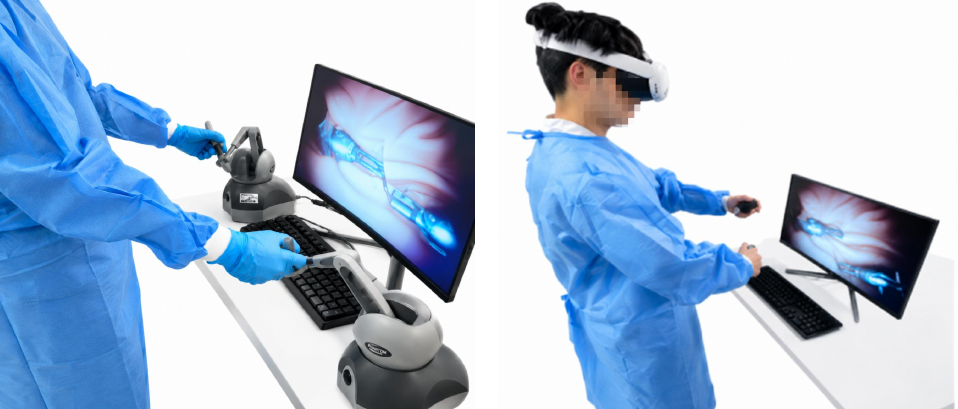}
  \caption{Operator-driven SurgCast with a physical 6-DoF surgical
  master device (left) and VR controllers (right) as sources of
  prospective action commands. Displays show the generated surgical view;
  action-dependent outputs are compared in Fig.~\ref{fig:action_branching}.}
  \label{fig:physical_system_demo}
\end{figure}

\begin{figure}[!tb]
  \centering
  \includegraphics[width=\columnwidth,pagebox=cropbox]{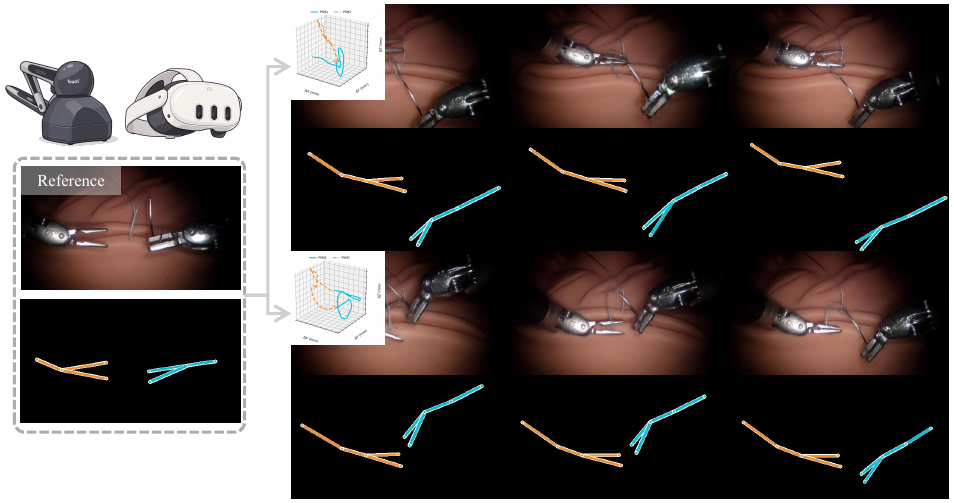}
  \caption{Action branching from the same initial image and skeleton
  (left). Distinct 6-DoF commands (insets) produce different predicted
  skeletons and video rollouts (upper and lower rows).}
  \label{fig:action_branching}
\end{figure}

Figure~\ref{fig:action_branching} qualitatively tests action responsiveness
from a fixed initial image and instrument skeleton. Distinct
6-DoF command trajectories produce different predicted skeletons
and generated instrument motions. The commands can be supplied
through a physical surgical master device or a companion VR
interface (Fig.~\ref{fig:physical_system_demo}).

Across 10 four-step, 73-frame rollouts, the causal student achieves
10.36 FPS in rollout throughput with 8.703\,GB peak GPU memory,
compared with 10.57 FPS
and 8.481\,GB for the original student.

\section{Conclusion}
SurgCast establishes predicted instrument structure as an explicit
interface between prospective robot actions and video generation.
Its two injection paths expose the same forecast as image-aligned
geometry and instrument-meaningful motion descriptors. Across the
evaluated datasets, this interface improves video fidelity and instrument
consistency over direct-action baselines. Ablations show that future
commands improve structural forecasting and that both injection paths
contribute under predicted skeletons; their combination performs best
across the six SutureBot ablation metrics. The gains also persist in
four-step causal generation. Together, these findings support forecasting
articulated structure and explicitly using its geometry and motion
semantics for surgical video control.

The evaluation uses supplied action windows; sustained closed-loop robot
control remains to be validated. Testing the interface under continuous
feedback is the next step toward robotic decision-making use.

\bibliographystyle{IEEEtranBST/IEEEtran}
\bibliography{references}

\end{document}